\documentclass[10pt, a4paper]{article}
\usepackage{swedish_w2v}
\usepackage{multibib}
\newcites{languageresource}{Language Resources}
\usepackage{graphicx}
\usepackage{tabularx}
\usepackage{soul}

\usepackage{epstopdf}
\usepackage[utf8]{inputenc}
\usepackage{MnSymbol,wasysym}
\usepackage{hyperref}
\usepackage{xstring}

\usepackage{color}

\title{ \vspace*{.5\baselineskip} \textbf{Hearing voices at the National Library - \\a speech corpus and acoustic model for the Swedish language}}

\name{Martin Malmsten, Chris Haffenden, Love Börjeson }

\address{KBLab, National Library of Sweden \\
         Humlegården, Stockholm \\
         www.kb.se/kb-labb\\
         \{martin.malmsten, chris.haffenden, love.borjeson\}@kb.se\\}
\abstract{
This paper details our work in developing new acoustic models for automated speech recognition (ASR) at KBLab, the infrastructure for data-driven research at the National Library of Sweden (KB). We evaluate different approaches for a viable speech-to-text pipeline for audiovisual resources in Swedish, using the wav2vec 2.0 architecture in combination with speech corpuses created from KB’s collections. These approaches include pretraining an acoustic model for Swedish from the ground up, and fine-tuning existing monolingual and multilingual models. The collections-based corpuses we use have been sampled from millions of hours of speech, with a conscious attempt to balance regional dialects to produce a more representative, and thus more democratic, model. The acoustic model this enabled, “VoxRex”, outperforms existing models for Swedish ASR. We also evaluate combining this model with various pretrained language models, which further enhanced performance. We conclude by highlighting the potential of such technology for cultural heritage institutions with vast collections of previously unlabelled audiovisual data. Our models are released for further exploration and research here: https://huggingface.co/KBLab. \\ \newline  }

\begin{document}

\maketitleabstract

\section{Introduction}

The emergence of unsupervised learning has had a major impact on the pace and scope of current AI development. Where the training of new and better models once demanded large amounts of labelled data, these can now be produced using principally unlabelled data, with just a fraction of the annotation previously required. This has led to significantly improved performance in natural language processing tasks for text via the widespread implementation of transformer-based BERT models \cite{devlin_bert_2019}. It has also recently been applied to audio data, with the use of similar architectures enabling comparable breakthroughs in speech recognition technology. Facebook (now Meta) AI thus released the wav2vec 2.0 model for self-supervised and the wav2vec U model for unsupervised learning of speech representations, which together promise a considerable expansion in automated speech recognition (ASR) technologies for lesser-resourced languages \cite{baevski_wav2vec_2020,baevski_unsupervised_2021}. 

However, this expansion will invariably reflect hierarchies of power and resources among the world’s languages. On the one hand, these new acoustic models have the potential to make speech technology available for a far greater range of languages, insofar as they reduce the bottleneck of a lack of labelled training data. To this end, Facebook (now Meta) AI has even released a cross-lingual model of speech recognition that was pretrained on 53 lesser-resourced languages, XLSR-53 \cite{conneau_unsupervised_2020}. Yet on the other hand, it is still likely to reinforce a chasm between the few high-resource languages that predominate and the rest. If speech technologies for languages like English will prove cutting edge, those for lesser-resourced languages risk remaining at a considerable remove from what is state-of-the-art. This is largely because of access to high quality training data: though pretraining a wav2vec model avoids the need for transcribed data, it still demands fairly significant levels of unlabelled audio data. For languages where this is not widely available, and where commercial actors perceive little incentive to invest in these resources, who will provide such data? 

In this context, national libraries and other cultural heritage institutions with large holdings of audiovisual material have a key role to play. By using their archives of high quality, language-specific data, such institutions can contribute towards a form of AI development with broader social benefits than that driven solely by private sector actors and big tech, especially for lower-resourced languages \cite{haffenden_making_2022}. This essentially democratic perspective informs our work with the training of new acoustic models for Swedish at KBLab at the National Library of Sweden (Kungliga Biblioteket, hereafter KB). As a result of having merged with the Swedish National Archive of Recorded Sound and Moving Images (Statens ljud- och bildarkiv, SLBA) in 2009, KB has a vast collection of (unlabelled) audio data, including national and local radio programmes. In turning to this data for acoustic modelling, we are building upon our earlier work for KB-BERT, which sought to produce a representative language model that corresponded to the “living language of the national community” \cite{malmsten_playing_2020}. Where we previously used the library’s collections to produce new possibilities for Swedish text processing, we are now doing the same for speech. Our work here thus forms part of the wider project of democratizing data value, while underpinning the development of a national AI infrastructure for Swedish, that we are engaged in at KBLab. 

This paper makes several contributions. Firstly, we explain how we used KB’s audiovisual collections to create a speech corpus and training data for a new generation of Swedish acoustic models. Secondly, we present the results of our evaluation process, where we compare the performance of our model, VoxRex, with existing models for Swedish ASR, including the monolingual VoxPopuli-sv and the multilingual XLSR \cite{wang_voxpopuli_2021,conneau_unsupervised_2020}. Thirdly, we highlight areas for further research, while also pointing towards the dramatic potential of such models for cultural heritage institutions with large collections of previously unlabelled audiovisual data. 


\section{ A Swedish speech corpus: P4 }

To pretrain any large model - e.g Wav2vec2 - thousands of hours of unlabelled speech is needed. VoxPopuli, a Wav2Vec2-based model trained by Meta, for example uses audio from the European parliament and XLSR uses Common Voice, BABEL and MultiLingual LibriSpeech. KB has multiple collections containing speech, such as movies, podcasts, audiobooks and radio and TV broadcasts that date back to the 1960s and 70s. In this work we have opted for a mix of local public radio, podcasts and audiobooks, with an emphasis on the former to maximize the number of speakers, types of speech and dialects, and thereby create a more representative and democratic dataset.

\subsection{ The P4 speech corpus }

Local public radio in Sweden, “P4”, currently consists of roughly 25 radio stations, many of which have existed since the 1970s. Through the legal deposit legislation, KB has physical copies of these broadcasts. Starting in the early ‘00s, the Swedish National Archive of Recorded Sound and Moving Images, SLBA, began digitizing its collection of local public radio broadcasts. This was done with the understanding that there would be no way to manually catalogue or categorize the material, but with the assumption that it could be made searchable or analyzed through automatic methods at some point in the future. The project is ongoing and has now digitized more than 2.7 million hours. No distinction is made between type of content, which means that music, news and sport reports, talk radio, people calling in, on-site reporting, etc. are digitized and stored together in audio files that are split according to a predetermined time boundary.

We used this data to create the P4 speech corpus. Only files from the last twenty years were selected.

\begin{table}[!ht]
\small
\begin{center}
\begin{tabularx}
    {\columnwidth}{X r ||X r ||X r}
    \textbf{year} & \textbf{\# files} & \textbf{year} & \textbf{\# files} & \textbf{year} & \textbf{\# files} \\ \hline
    \textbf{2002} & 25295 & \textbf{2003} & 193574 & \textbf{2004} & 224736 \\ 
    \textbf{2005} & 225709 & \textbf{2006} & 243918 & \textbf{2007} & 259763 \\
    \textbf{2008} & 259806 & \textbf{2009} & 259982 & \textbf{2010} & 266997 \\
    \textbf{2011} & 272858 & \textbf{2012} & 308760 & \textbf{2013} & 449716 \\
    \textbf{2014} & 456860 & \textbf{2015} & 455665 & \textbf{2016} & 456759 \\
    \textbf{2017} & 455460 & \textbf{2018} & 455541 & \textbf{2019} & 458422 \\
    \textbf{2020} & 474282 & \textbf{2021} & 390536 \\
\end{tabularx}
\caption{Broadcast year distribution}
\end{center}
\end{table}

One of the main reasons for choosing specifically local public radio was to get a more diverse corpus in terms of dialects and types of speech. The assumption is that this will improve downstream tasks where the speaker does not speak “standard Swedish” and/or read from a manuscript. 

\begin{table}[!ht]
\small
\begin{center}
\begin{tabularx}
    {\columnwidth}{X r ||X r}
    \textbf{channel} & \textbf{\# files} & \textbf{channel} & \textbf{\# files} \\ \hline
    P4 Blekinge & 247993 & P4 Jönköping & 258820 \\
    P4 Norrbotten & 262690 & P4 Västerbotten & 246940 \\
    P4 Plus* & 11113 & P4 Dalarna & 262158 \\
    P4 Jämtland & 234806 & P4 Örebro & 237203 \\
    P4 Västmanland & 251748 & P4 Riks & 273309 \\
    P4 Gävle & 253933 & P4 Kalmar & 253555 \\
    P4 Skaraborg & 239971 & P4 Västernorrland & 245273 \\
    P4 Södertälje & 34234 & P4 Göteborg & 248897 \\
    P4 Kristianstad & 256075 & P4 Sörmland & 250144 \\
    P4 Värmland & 250247 & P4 Sjuhärad & 240130 \\
    P4 Gotland & 237801 & P4 Kronoberg & 255959 \\
    P4 Stockholm & 264287 & P4 Västmanland & 264119 \\
    P4 Halland & 256332 & P4 Malmö & 248182 \\
    P4 Uppland & 247107 & P4 Östergötland & 261613 \\
\end{tabularx}
\caption{Regional distribution}
\end{center}
\end{table}

\subsection{Speech detection and extraction}
The main preprocessing step is to identify parts of the audio that is considered speech. Having a large collection to begin with, we have the luxury of being conservative in our selection, i.e. we do not have to find all speech, just enough for our training.

Audio files containing speech used for training were extracted using the following heuristic: each sound file is split into frames of 20ms length. Voice and silence detection is run on each frame. Silence is defined as audio that does not contain voice and has a dBFS of less than -40. We used Silero VAD to detect voices with the level parameter set to 2 \cite{Silero}. Frames are then bundled into chunks of 50 frames, i.e 1000ms. A chunk is valid only if it contains voice or silence and their respective ratios are between certain thresholds. Information about consecutive valid chunks with a combined length of more than 30 seconds are written to disk. The result is a master file with pointers to those parts of a file that contain viable speech.

A corpus of arbitrary size can then be generated by randomly selecting 30 second spans of speech from the master file up to a predetermined total amount (1k, 10k, 100k, etc. hours) and writing these to disk with available metadata, i.e broadcast channel, filename and timestamp. Results show that roughly 50\% of the total running time is tagged as speech, making the estimated size of the complete corpus 1.4M hours with 100k hours added yearly.

The corpus currently exists in its rawest form with the only preparation made is detected speech and the only metadata is broadcast date and the original broadcast channel. Audio is downsampled to 16KHz mono.

\subsection{Other sources}
Alongside P4 other sources of speech were used in the training. Using the same speech detection heuristic as for P4, a total of 1100 hours of speech was extracted from audiobooks and podcasts.

\section{A Swedish acoustic model: VoxRex}
We used this speech corpus and additional audio data to train multiple versions of \emph{VoxRex}, a Wav2Vec2 model with ~300 million parameters as first described by Facebook AI \cite{baevski_wav2vec_2020}. This corresponds in size to the original Wav2Vec2 Large. The number of attention heads, hidden layers and intermediate size was set to 16, 24 and 4096 respectively. Multiple versions were trained to compare the effect of more data and longer training time.

\subsection{Pretraining}
All training was done on a single NVIDIA DGX A100 with eight 40GB GPUs. The Fairseq library was used. Pretraining took approximately 21 days to reach 400k steps on 8 GPUs. The number of gradient accumulation steps was set to eight to simulate a world size of 64 GPUs. The max learning rate was set to 5e-3.

\begin{table}[h!]
\small
\begin{center}
\tabcolsep=0.11cm
\begin{tabularx}
    {\columnwidth}{X r r r}
    \textbf{name} & \textbf{updates} & \textbf{hours of speech} & \textbf{corpus} \\ \hline
    VoxRex-A & 200k & 1000 & P4-1k \\
    VoxRex-B & 200k & 2100 & P4-1k++ \\
    VoxRex-C-200k & 200k & 11100 & P4-10k++ \\
    VoxRex-C & 400k & 11100 & P4-10k++ \\
\end{tabularx}
\caption{VoxRex versions, training time and amount of data}
\end{center}
\end{table}

\section{Evaluation}
To evaluate these various VoxRex versions and compare them with other models, we fine-tuned for an ASR task and measured WER on a subset held back from the training data, containing distinct sentences not repeated by other speakers in the training set. We evaluate VoxRex against two models that can be used for Swedish downstream tasks, which were released during the training period: a monolingual model, VoxPopuli-sv, and a large multilingual model, XLSR \cite{wang_voxpopuli_2021,conneau_unsupervised_2020}.

\subsection{Labelled datasets}
We use two labelled datasets: NST and CommonVoice 6.1 \cite{birkenes_2020,commonvoice:2020}. NST is a dataset created by Nordisk språkteknologi holding
AS and donated to Språkbanken at the National Library of Norway. Common Voice 6.1 contains 6349 unique sentences spoken by 222 persons,
in total 12 hours. NST contains ~250k unique sentences spoken by 1000 persons.

\subsection{ASR fine-tuning task}
The Fairseq library was used for fine-tuning all the CTC. Huggingface \cite{wolf-etal-2020-transformers} was used to train the encoder-decoder model.

We fine-tuned all the models for 120k updates on a NST + CommonVoice 6.1 dataset. The evaluation was carried out using a test set containing 2\% of total sentences. The CommonVoice part of the test set corresponds to the one used by Huggingface during the XLSR Fine-tuning Week event.

\begin{figure}[h!]
    \centering
    \def\svgwidth{\columnwidth}
    \scalebox{1.0}{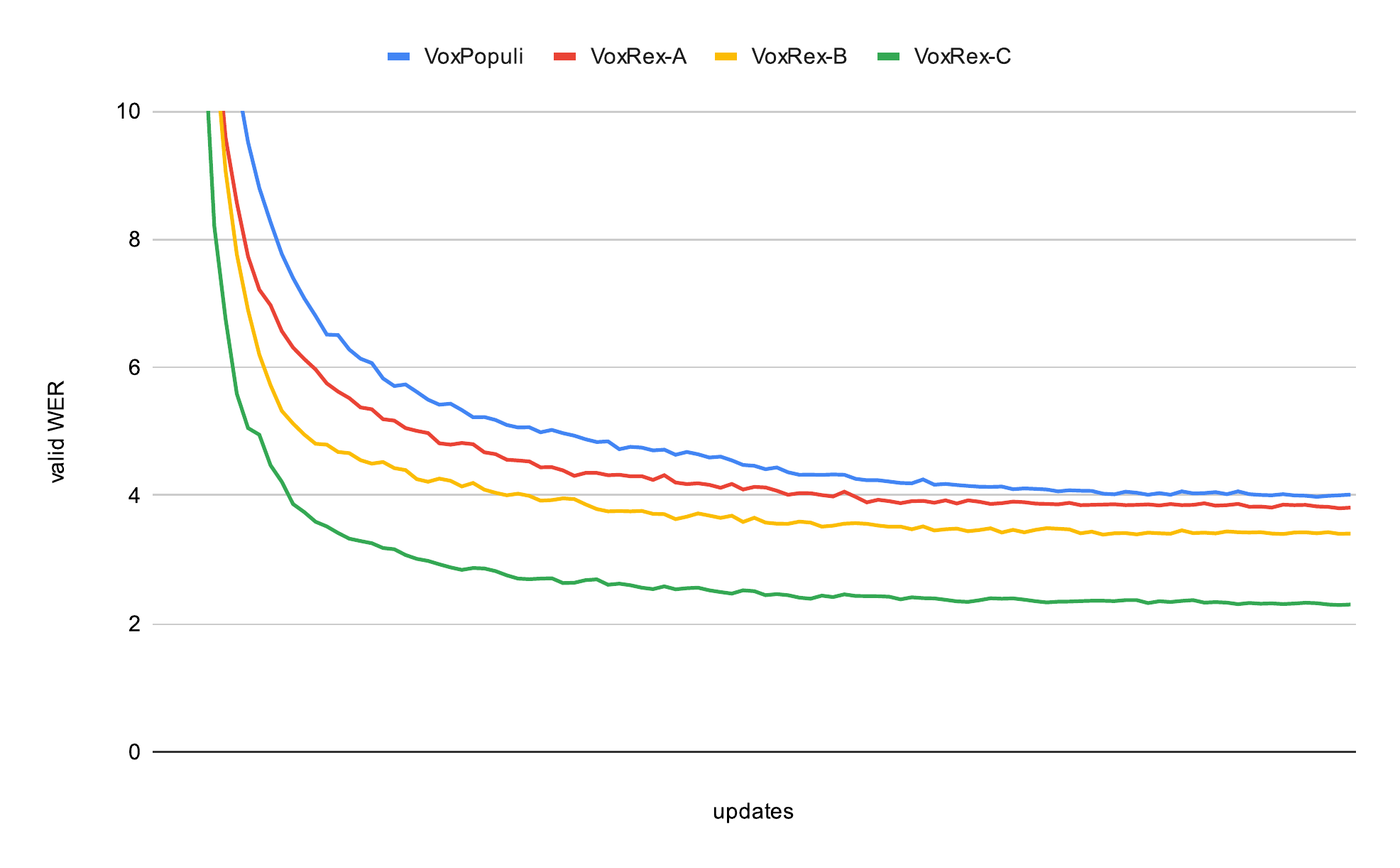}
\caption{Valid WER during finetuning}
\end{figure}

Results show that VoxRex outperforms both XLSR and VoxPopuli. The VoxRex-C version reaches the same WER as VoxPopuli after just 15\% of training. 

\begin{table}[h!]
\small
\begin{center}
\tabcolsep=0.11cm
\begin{tabularx}
    {\columnwidth}{X r r r r r r}
    \textbf{} & \multicolumn{2}{c}{\textbf{raw}} & \multicolumn{2}{c}{\textbf{social}} & \multicolumn{2}{c}{\textbf{gov}} \\
    \cline{2-7}
    \textbf{model} & NST & CV & NST & CV & NST & CV \\ \hline
    XLSR & 7.15 & 17.99 & 14.47 & 11.74 & 11.35 & 15.25 \\
    VoxPopuli & 4.21 & 15.12 & 13.37 & 11.48 & 10.09 & 14.87 \\
    VoxRex-A & 3.88 & 13.05 & \textbf{12.12} & 9.85 & 9.73 & 12.88 \\
    VoxRex-B & 3.40 & 10.72 & 12.59 & 8.71 & 9.8 & 11.4 \\
    VoxRex-C & \textbf{2.5} & \textbf{8.49} & 12.69 & \textbf{7.37} & \textbf{9.15} & \textbf{9.29} \\
    VoxRex-C-BART & 3.5 & 10.39 & n/a & n/a &n n/a & n/a \\
\end{tabularx}
\caption{Evaluation with and without LM}
\end{center}
\end{table}

\subsection{Evaluation with language models}
We create two 4-gram models using KenLM: “4-gram social” and “4-gram gov” based on internet forums and government publications respectively. It is clear from the evaluation that the basis for the language model can have a large impact on performance. The social model consistently improves WER on the CV test set - compared to raw CTC output without a model - for all models, while the gov model reduces performance for VoxRex-B and -C. On the other hand, the gov language model performs better than the social one on the NST test set for all acoustic models, though results without any language model outperforms both.

\subsection{Encoder-decoder model}
A different approach to using the Wav2Vec2 encoder with CTC is an encoder-decoder model where the encoder is coupled with a text decoder. This has been explored by many, e.g \cite{DBLP:journals/corr/ChanJLV15} using recurrent networks and more recently using transformer based networks in \cite{DBLP:journals/corr/abs-1907-12461}. Being sequence-to-sequence models they can be used both for automatic speech recognition (ASR) and speech translation (ST) tasks.

Training such models purely on labeled data would however require vast amounts of transcribed speech. Another approach detailed in \cite{DBLP:journals/corr/abs-2104-06678} is to warm-start both the encoder and decoder with existing checkpoints before training. In theory, this would use the speech understanding of the acoustic model in conjunction with the the language understanding of the text model to transcribe speech gained during unsupervised pre-training. Models suitable for this approach are for example GPT and the decoder part of BART and T5. The National Library has previously pre-trained a Swedish BART using the same data as it's BERT model \cite{lewis_bart_2019,kurtz_2021,malmsten_playing_2020}. Consequently the VoxRex-C-BART model was initialized using the BART decoder and the Wav2Vec2 encoder and fine-tuned using the same labeled data as the other models.

Generally, it seems that this method works more like translation than transcription in the sense that order plays a lesser role. The model chooses words that make sense from a language perspective, though that does not necessarily match the audio very well. It also shows the same tendency as generative models to sometimes get stuck in a loop. 

The following output from the validation exemplifies this:

\texttt{mismatch: En purpur nektarin. -> En purpur aprik aprik aprik aprikos.}

“En purpur nektarin” (a purple nectarine) and “En purpur aprikos” (a purple apricot) are sentences about similar fruits, which might imply that they are used in similar contexts in text, but have very different pronunciation. In other instances, long audio files with multiple sentences get truncated to one sentence.

Even though the results were poor compared to CTC the method is still of interest since it can handle sequence-to-sequence tasks, e.g punctuation restoration, capitalization and correction, as a part of transcription.

\subsection{External evaluation}
External evaluation of the model has been carried out in \cite{lagerlof_swedish_2022}, which compared VoxRex-B to Google’s speech-to-text API. The results show that VoxRex-B outperformed Google by roughly ten percentage points on a test set of news broadcasts from Swedish radio, 29.4\% WER compared to 38.7\%, even though VoxRex (unsurprisingly) performed poorly on the English parts of the test.

\subsection{Regional differences}
The NST data has information about the speaker’s region of birth and “region of youth”. Although this does not necessarily imply each speaker’s dialect, we use it as a proxy on an aggregate level. The test set does not include enough data to evaluate reliably by region, so evaluation is done on the whole training set. While this does not provide accurate numbers on WER per region, it can be used to compare the models. The table below shows that performance is improved for every dialect and that some dialects are harder than others to transcribe, with the south-west (“Västra sydsverige”) being the most challenging dialect.

\begin{table}[h!]
\small
\begin{center}
\tabcolsep=0.11cm
\begin{tabularx}
    {\columnwidth}{Xlllll}
    \textbf{region / model} & \textbf{XLSR} & \textbf{VoxPopuli} & \textbf{A}  & \textbf{B} & \textbf{C} \\ \hline
    Dalarna & 5.62 & 3.17 & 2.74 & 2.74 & 1.67 \\
    Göteborg w. env. & 5.64 & 3.26 & 2.86 & 2.82 & 1.78 \\
    Mellansverige & 5.62 & 3.30 & 2.85 & 2.83 & 1.79 \\
    Norrland & 6.27 & 3.68 & 3.12 & 3.20 & 1.95 \\
    Stockholm w. env. & 4.57 & 2.64  & 2.26 & 2.30 & 1.43 \\
    Västergötland & 5.53 & 3.26 & 2.78 & 2.81 & 1.76 \\
    Västra sydsverige  & 7.62 & 4.25 & 3.77 & 3.84 & 2.1 \\
    Västsverige & 5.40 & 3.16 & 2.65 & 2.71 & 1.6 \\
    Östergötland & 5.65 & 3.23 & 2.81 & 2.77 & 1.68 \\
    Östra sydsverige & 6.68 & 3.80 & 3.29 & 3.23 & 1.94 \\
\end{tabularx}
\caption{Pseudo-WER per region and model}
\end{center}
\end{table}

As shown above, all versions of VoxRex outperform the multilingual XLSR and monolingual VoxPopuli with VoxRex-C being clearly better than the other versions. However, even though VoxRex-C proved better at apparently difficult dialects, the relation between the democratic (i.e. regionally diverse) data in our training corpus and this improved performance requires further investigation before direct causality can be established. 

\section{Conclusion}
In this paper, we have described how we used KB’s audiovisual data to produce a new acoustic model for Swedish. We have demonstrated how VoxRex, our monolingual model trained on the P4 corpus, podcasts and audiobooks, outperformed existing multilingual and monolingual models on a speech-to-text task, even though only rudimentary processing was carried out to detect viable speech. As with our previous work in making a Swedish BERT for text analysis, this once again shows the value of high quality, language-specific data to continued AI development, especially for smaller languages like Swedish. 

Two broader conclusions can be drawn from this particular work. Firstly, the emergence of unsupervised learning creates a special role for cultural heritage institutions with large digitized collections in building new AI infrastructures—particularly for lesser-resourced languages. On the one hand, these institutions can contribute to a significant democratization of AI tools. With sufficient computational resources and data science expertise located within such an institution — as is the case with KBLab at KB — new acoustic models can be trained that lay the groundwork for smaller organizations to fine-tune models for other tasks. Since fine-tuning involves considerably less resources, this makes cutting edge performance available for actors not usually invested in AI development or scholarly work in the field. On the other hand, these models also promise important benefits for the cultural heritage institutions themselves. The presence of a state-of-the-art model for ASR could lead to the mass transcription of millions of hours of hitherto unlabelled speech data, transforming audiovisual collections that were previously largely unsearchable due to lack of metadata into exciting new sites of research. In this sense, it is a symbiotic relationship: cultural heritage data enables AI innovation, which in turn makes possible novel exploration, and enrichment, of digitized cultural heritage.

The second general conclusion relates to the character of the digitization process. Namely, that the digitization of material at cultural heritage institutions need not necessarily presume the creation of perfect descriptions, or being manually labelled, in order to prove valuable. Such descriptions contribute nothing to the pretraining phase, and contents-based descriptions can be generated at a later stage rather than painstakingly catalogued. In short, what this example of library-based AI development suggests is that a “digitize first, ask questions later” approach seems more viable than ever, especially in terms of being able to reap the unintended and unforeseeable benefits created from using such digital holdings. 

\subsection{Further work}
The P4 corpus is largely unexplored when it comes to the actual content of the material. The geographical location of the station gives some indication of the prevalent dialect, but this is in no way guaranteed. Furthermore, there are central broadcasts such as news broadcasts that are the same for every channel. In short, the amount of actual dialect spoken is unclear.

However, with the VoxRex model trained it is now possible to move on to downstream tasks such as dialect classification, sentiment analysis, etc. This can in turn be used to further enhance the P4 corpus with more fine-grained labels. We are in essence bootstrapping further work with this first model. The major unanswered question, though, is the extent to which even more data or training (or both) will provide yet better results. This work is ongoing at KBLab.

\section{Availability of corpus and models}
We are making VoxRex and the fine-tuned versions freely available with a CC0 license via Huggingface.\footnote{https://huggingface.co/KBLab} The P4 corpus cannot be freely distributed, due to copyright restrictions, but it is available to be used by researchers in-situ at the KBLab in Stockholm.

\section{Acknowledgements}
Even though the digitized material was not usable at the time, the decision to “blindly” digitize local public radio broadcasts has been crucially important for this project. We wish to thank everyone involved with this, especially those responsible at what was then the Swedish National Archive of Recorded Sound and Moving Images (SLBA) and is now part of the National Library of Sweden.

The freely-available resources from both CommonVoice and NST were instrumental in reaching high performance for the speech-to-text task.

We have used open source software from Facebook and Huggingface extensively.

This work was done with the support of the National Library of Sweden in general and KBLab in particular.

\section{Bibliographical References}\label{reference}

\bibliographystyle{swedish_w2v}
\bibliography{swedish_w2v}      

\end{document}